\documentclass[12pt]{article}
\usepackage{amsmath}
\usepackage{graphicx,psfrag,epsf}
\usepackage{enumerate}
\usepackage{natbib}
\usepackage{float}
\usepackage{siunitx}
\usepackage{amsthm}
\usepackage{amssymb}
\usepackage{amsfonts}
\usepackage[title]{appendix}

\usepackage{color}
\usepackage{caption, subcaption}
\usepackage{numprint}
\npthousandsep{,}

\usepackage{algorithm}
\usepackage[noend]{algpseudocode}

\newcommand{\blind}{0}
\newcommand{\xbs}[1]{\boldsymbol{x}_{\setminus #1}}

\newcommand{\defeq}{\overset{\text{\tiny def}}{=}}

\newtheorem{theorem}{Theorem}[section]

\begin{document}
\date{}

\def\spacingset#1{\renewcommand{\baselinestretch}%
{#1}\small\normalsize} \spacingset{1}


\if0\blind
{
  \title{\bf Interpretable Architecture Neural Networks for Function Visualization}
  \author{Shengtong Zhang\hspace{.2cm}\\
    Department of Industrial Engineering and Management Sciences, \\ Northwestern University\\
    and \\
    Daniel W. Apley \\
    Department of Industrial Engineering and Management Sciences,\\ Northwestern University\\}
  \maketitle
} \fi

\if1\blind
{
  \bigskip
  \bigskip
  \bigskip
  \begin{center}
    {\LARGE\bf Title}
\end{center}
  \medskip
} \fi

\bigskip
\begin{abstract}
In many scientific research fields, understanding and visualizing a black-box function in terms of the effects of all the input variables is of great importance. Existing visualization tools do not allow one to visualize the effects of all the input variables simultaneously. Although one can select one or two of the input variables to visualize via a 2D or 3D plot while holding other variables fixed, this presents an oversimplified and incomplete picture of the model. To overcome this shortcoming, we present a new visualization approach using an interpretable architecture neural network (IANN) to visualize the effects of all the input variables directly and simultaneously. We propose two interpretable structures, each of which can be conveniently represented by a specific IANN, and we discuss a number of possible extensions. We also provide a Python package to implement our proposed method. The supplemental materials are available online.
\end{abstract}

\noindent%
{\it Keywords:} Function Visualization;  Interpretable Machine Learning; Neural Network
\vfill

\newpage
\spacingset{1.75} 

\section{Introduction}

Scientists and engineers often want to visualize the output of a black box function in order to understand and interpret the effects of input variables on the output (aka response) variable. For example, most surrogate models of complex computer simulations are notorious for their black box property and lack interpretability. Visualizing a black box model helps to understand how the inputs affect the predicted response variable, whether the predictive relationships seem intuitively reasonable, and how decisions can be made based on the model.

We denote the black box function by $f(\boldsymbol{x})$, where $\boldsymbol{x} = (x_1, x_2, \cdots, x_d)$ is a $d$-dimensional vector of input variables, and $f(\boldsymbol{x})$ is the scalar response. $f(\boldsymbol{x})$, as a function of $\boldsymbol{x}$, is often referred to as a response surface. In reality, many functions estimated from a large data set or a complex computer simulation model have many input variables, which makes visualization of the joint effects of all the input variables challenging. 

Of the existing works that aim to interpret black box models, perhaps the most common and the most closely related works are ``partial dependence" (PD) plots \citep{friedman2001greedy} and ``individual conditional expectation" (ICE) plots \citep{goldstein2015peeking}. Each PD plot shows the marginal effect of one or two selected input variables by integrating the response variable over the marginal distribution of the omitted input variables, whereas each ICE plot displays a collection of curves that are functions of a single selected input variable of interest, one curve for each fixed combination of the $d-1$ omitted input variables. Accumulated local effects (ALE) plots \citep{apley2020visualizing} improve upon PD plots by offering much faster computations and more accurate results when the input variables are highly correlated. Closely related to ICE plots, trellis plots \citep{becker1996visual, becker1994trellis} are a series of plots of the response variable as a function of a pair of selected input variables with the omitted variable(s) held fixed, with a separate plot for each fixed combination of the omitted variables. Although trellis plots typically present a clear and fairly complete picture of $f(\boldsymbol{x})$ for the case of $d = 3$, for $d > 3$ they become cumbersome, since there are too many fixed combinations of the $d-2$ omitted variables to consider. Whereas PD, ICE, trellis and ALE plots focus on visualizing the effects of one or two variables with each plot, and do not present a clear picture of the interactions between the selected and omitted input variables for large $d$, our approach aims to visualize the joint effects of all input variables. Moreover, \citet{agarwal2021neural} combined deep neural networks with generalized additive models to increase interpretablity. The primary distinction is that to enable easy visualization, the neural network architecture in \citep{agarwal2021neural} is restricted to additive functional relationships of the form $f(\boldsymbol{x}) = f_1(x_1) + f_2(x_2) + \cdots + f_d(x_d)$ and, therefore, cannot be used to represent and visualize interactions between the input variables. 

Other less closely related works focus on calculating a variable importance measure for each input variable. \citet{breiman2001random} introduced the idea of permutation-based feature importance in random forest models, and it was later extended to general black box models \citep{fisher2019all, konig2021relative}. Based on game theory concepts, \citet{lundberg2017unified} used Shapley values to compute feature importance. However, the feature importance measures only provide a scalar numerical score of the importance of each input variable, without revealing how the variables affect the response variable.

Instead of singling out the effects of the original variables/features, some approaches aim to visualize the topological structure of $f(\boldsymbol{x})$. These methods largely focus on identifying the number and locations of local minima and maxima of $f$ and, subsequently, on identifying paths in the input space to traverse between the minima and maxima. \citet{gerber2010visual} used the Morse-Smale complex to segment the $d$-dimensional input variable space into several non-overlapping regions over which $f$ is monotonic and located local minima and maxima of $f$. To visualize the topology of $f$ over each segmented region, they constructed certain regression surfaces and embedded them in 2D as a simplified representation of $f$ in each region. This approach was applied to nuclear reactor safety analysis and visualization \citep{maljovec2013exploration}. \citet{harvey2012enhanced} further used the Reeb graph to shatter the loops created by the Morse-Smale complex and provide a topologically simpler visualization. Although the local effects of input variables can be interpreted from the embedded regression curve for each segment in the Morse-Smale complex, the global effects are difficult to interpret, especially as the number of segments increases. Moreover, the method is not well suited for visualizing global interactions between a set of inputs.

Our visualization approach is based on the following approximate representation of $f$: 
\begin{equation}
    f(x_1, x_2, \cdots, x_d) \approx g(x_j, h(\xbs{j})) \quad \mbox{for some } j \in \{1,2,\cdots, d\}.
    \label{function}
\end{equation}
The notation ``$\xbs{j}$'' represents all the input variables excluding $x_j$. For simplicity, we focus on functions $f$ defined on the Cartesian product  $\prod \limits_{j=1}^d I_j$, where each $I_j \in \mathbb{R}$ is a closed interval.  In Section \ref{sec:first_level} we show that for each $j \in \{1,2,\cdots, d\}$ any continuous $f$ on $[0,1]^d$ can be arbitrarily closely approximated by the structure (\ref{function}) for some continuous functions $g$ and $h$ (although for some $j$ the resulting functions $g$ and $h$ will be better behaved than for other $j$). We also describe an approach for estimating $g$ and $h$ and for selecting the index $j$ that leads to a good approximation.

\begin{figure}[H]
\centering
\includegraphics[width=\textwidth]{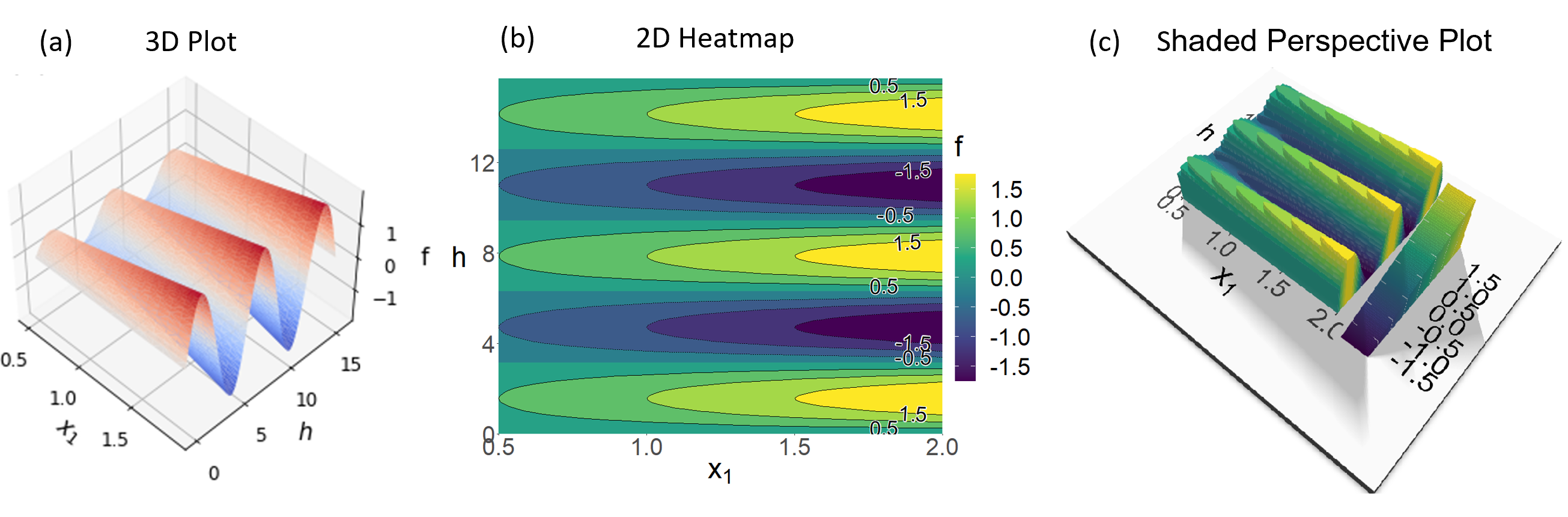}
\caption{An example of a visualization per structure (\ref{function}) of the harmonic wave function $f$ with (a) a 3D plot, (b) a 2D heatmap of the same function and (c) a shaded perspective plot. See the online version for color figures.}
\label{fig:harmonic_intro}
\end{figure}

For visualizing $f$ via (\ref{function}) with a particular input variable $x_j$ singled out, one can construct a 3D plot of $f$ vs $x_j$ and $h$. To illustrate, consider the harmonic wave function from physics
\begin{equation}
    f(\boldsymbol{x}) = x_1 * \sin [(2 \pi / x_2) *x_3+x_4].
    \label{eqn:harmonic2}
\end{equation}
Here $f$ is the displacement of a given point on the wave, $x_1 \in [0.5,2]$ is the amplitude, $x_2 \in [0.5,2]$ is the wavelength, $x_3 \in [0, 1]$ is the position of that point (e.g., the distance from the source of the wave), and $x_4 \in [0, \pi]$ is the phase of the wave. $f$ can be represented as (\ref{function}) with $x_1$ singled out, $f(\boldsymbol{x}) = x_1 * \sin[h(\xbs{1})]$, and $h(\xbs{1}) = (2 \pi / x_2) *x_3+x_4$. The corresponding 3D visualization plot ($f$ vs $x_1$ and $h$) is shown in Figure \ref{fig:harmonic_intro}(a). Such a plot allows one to visualize the manner in which $f$ depends on the single variable $x_j$, including how $x_j$ interacts with some latent function ($h$) that captures the full extent of the $x_j$ interactions with the other $d-1$ input variables. 

In this paper, we visualize the estimated functions mainly using static 3D plots. However, considering that static 3D plots may be easily misinterpreted due to visual misperception \citep{cleveland1984graphical}, we recommend that users plot the functions using whatever visual rendering they prefer, e.g., using software that allows one to interactively rotate the 3D plots to better avoid misinterpreting or missing features of the plot, supplementing the 3D plot with a 2D heat map with contour lines added, and/or using advanced shading and perspectives. In Figure \ref{fig:harmonic_intro}, we also include a 2D heatmap of the same function and a shaded perspective plot using the software \textit{rayshader} \citep{rayshader}.

In order to understand how $x_j$ interacts with the other individual input variables and, more generally, to understand the joint effect of all $d$ input variables on $f$, our approach proceeds hierarchically and approximates $h$ with a structure analogous to (\ref{function}) with a second input variable singled out, and so on. We describe this approach in Section \ref{sec:ovh} and refer to it as the ``original variable hierarchical" (OVH) structure. Moreover, we also consider a more general structure that assumes the hierarchical $h$ functions are functions of certain linear combinations of the original inputs that can further enhance visualization. We refer to this as the ``disjoint active subspace hierarchical" (DASH) structure and describe it in Section \ref{sec:dash}.

The remainder of the paper is organized as follows. In Section \ref{sec:first_level} we present a theorem to show that any continuous function $f$ can be arbitrarily closely approximated by a model with the more interpretable structure (\ref{function}), and we describe our \emph{interpretable architecture neural network (IANN)} model to estimate the functions $g$ and $h$ in (\ref{function}). In Section \ref{sec:hierarchical}, we develop the OVH and DASH structures to visualize and interpret the joint effects of all the variables by hierarchically decomposing the latent function $h$, both of which can be conveniently represented with a specific IANN model. We note that for visualizing complex black box simulation response surfaces that are expensive to evaluate, one should first fit a surrogate model \citep{sacks1989design, kennedy2000predicting, kennedy2001bayesian} to the simulation data and then use the surrogate model as $f$. In Section \ref{sec:algorithm} we present algorithms for finding the appropriate ordering of the input variables in the hierarchical decompositions. In Section \ref{sec:numerical}, we provide additional numerical examples to illustrate function visualization using our IANN approach.  In Section \ref{sec:conclusion}, we discuss a number of potential extensions of the IANN approach. 

\section{IANN Structure for the First Hierarchical Level}
\label{sec:first_level}

This section describes how to use the IANN model to approximate $f$ by structure (\ref{function}) and how it facilitates the visualization of $f$.
In Section \ref{sec:first_level_approx}, we present Theorem \ref{thm} to show that the proposed structure (\ref{function}) can approximate any continuous function on $[0,1]^d$ for some continuous functions $g$ and $h$. To estimate $g$ and $h$, we introduce the IANN architecture to approximate $f$ for each singled out variable $x_j$ in Section \ref{sec:architecture}. In Section \ref{sec:ICE}, we compare and draw a connection between PD plots, ICE plots, and the IANN visualization plots to show that the latter has advantages that lead to clearer interpretation of $f$. This decomposition can be used as a stand-alone approach to visualize the effects of each input (as illustrated in Figure \ref{fig:harmonic_intro}), if it is repeated in a nonhierarchical manner for each input $x_j$ for $j = 1, 2, \cdots, d$. The main information that is missing in this visualization is the disentanglement of the joint effects of all inputs, which is the subject of Section \ref{sec:hierarchical}. For this, the decomposition in this section provides the basic building block.

\subsection{IANN approximation theorem}
\label{sec:first_level_approx}
The following theorem (see Appendix D for a proof)  guarantees that the interpretable structure we proposed in (\ref{function}) can approximate any continuous function $f$ on $[0,1]^d$. 
\begin{theorem}
Let $f: [0,1]^d \to \mathbb{R}$ be a continuous function. For any $\epsilon > 0$ and $j \in \{1,2,\cdots, d\}$, there exist continuous functions $g$ and $h$ satisfying: 
\begin{equation}
    \left|f(\boldsymbol{x}) - g(x_j, h(\xbs{j}))\right| < \epsilon, \quad \forall \boldsymbol{x} = (x_j, \xbs{j}) \in [0,1]^d.
    \label{eqn:approx}
\end{equation}
\label{thm}
\end{theorem}
\noindent More generally, the theorem above holds if $f$ is defined on the Cartesian product of intervals in each $x_j$. 

The significance of the theorem is that the structure (\ref{function}) can be used to approximate arbitrarily closely a continuous $f$, and this approximation can then be used to visualize $f$ as in Figure \ref{fig:harmonic_intro}. This visualization is related to ICE plots but provides more clarity in the sense that we discuss in Section \ref{sec:ICE}.  Moreover, since $h$ is continuous, one can apply Theorem \ref{thm} again to show that $h$ can be closely approximated by the same structure (\ref{function}), and this process can be repeated to yield the hierarchical decomposition described in Section \ref{sec:hierarchical}.

Notice that Theorem \ref{thm} applies for each $j \in \{1, 2, \cdots, d\}$, although there is a caveat that should be pointed out. For some $j$, the corresponding $h$ may be so complex that it cannot realistically be estimated and visualized in the subsequent levels. However, for all the real functions $f$ we have considered, for at least one of the input variables $x_j$ we were able to accurately approximate $f$ with an $h$ that was well behaved and could subsequently be approximated via the same structure (\ref{function}) in the next level. Moreover, our algorithm (in Section \ref{sec:algorithm}) automatically finds a good hierarchical ordering of the input variables to yield an accurate approximation of $f$, and the accuracy of the approximation can be easily quantified as a check to verify whether the function $f$ in question can indeed be approximated by an IANN structure.

\subsection{IANN architecture}
\label{sec:architecture}

\begin{figure}[H]
\includegraphics[width=\textwidth]{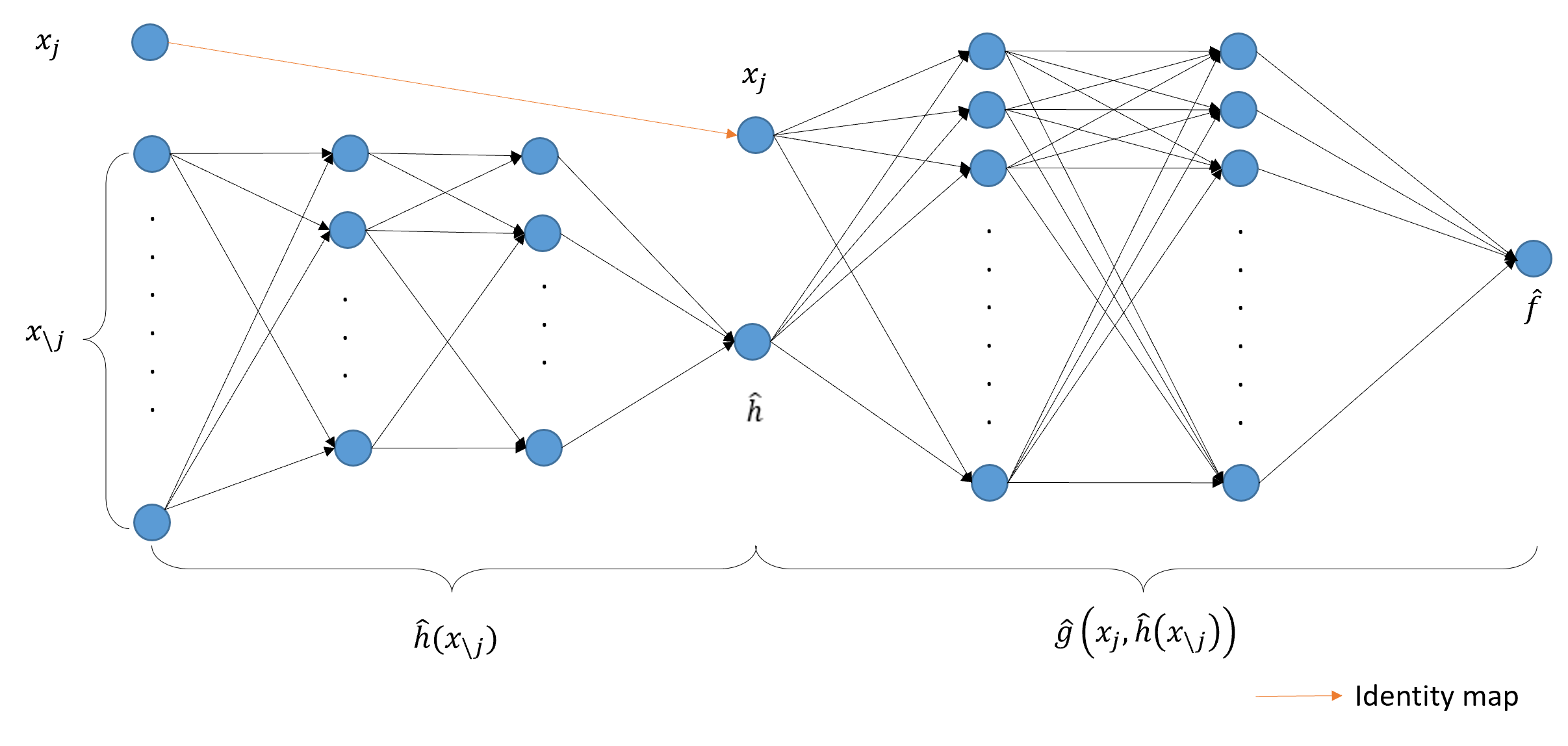}
    \caption{IANN architecture representing Eq. (\ref{function}) at the top hierarchical level. The  bottle-neck layer in the middle consists of only two nodes, which represents $x_j$ and $\Hat{h}(\xbs{j})$.}
\label{fig:exploration1}
\end{figure}

To estimate the functions $g$ and $h$, we use the customized neural network architecture depicted in Figure \ref{fig:exploration1}, which we refer to as an IANN.

The input layer has $d$ nodes that represent the $d$ input variables, and the single node in the output layer is the approximation of the original function $f(\boldsymbol{x})$. Due to Theorem \ref{thm} and the universal approximation theorem of neural networks \citep{hornik1989multilayer,cybenko1989approximation,hornik1991approximation}, with enough layers and nodes the IANN in Figure \ref{fig:exploration1} can approximate any continuous function $f$ by the structure  (\ref{function}) for some functions $g$ and $h$ estimated in the process of fitting the IANN model.

The bottle-neck layer in the middle has only two nodes, representing $x_j$ and $\Hat{h}$. $x_j$ is directly connected to the first input node by an identity map, and $\Hat{h}$ is the estimation of the latent function $h(\xbs{j})$ in Eq. (\ref{function}). In the layers to the left of the bottle-neck layer, we connect the $d-1$ nodes (which represent the remaining input variables $\xbs{j}$) to node $\Hat{h}$ by a fully-connected neural network denoted by $\Hat{h}(\xbs{j})$. In the layers to the right of the bottle-neck layer, we use another fully-connected network (denoted by $\Hat{g}$, with inputs $x_j$ and $\Hat{h}$) to represent the function $g$ as an approximation of $f(\boldsymbol{x})$.

Denote the training data input/response observations by $\{(x_1^i, x_2^i, \cdots, x_d^i, f^i), \; 1 \leq i \leq N\}$ and the IANN response prediction by $\Hat{f}^i = \Hat{g}(x_j^i, \Hat{h}(\boldsymbol{x}_{\setminus j}^i))$, where $N$ is the number of observations in the training set. The training data were generated using the customized Latin Hypercube sampling described in the Appendix B. Note that even when $f$ represents a simulation response surface that is expensive to evaluate, $N$ can be chosen quite large, because $f$ is first replaced by a surrogate model fit to the simulation data, and generating response observations from the surrogate model is inexpensive. To fit the IANN (i.e., estimate the weights and biases) to the training data, we use standard squared error loss:
\begin{gather}
    Loss = \frac{1}{N}\sum \limits_{i=1}^N \left|f^i - \Hat{f}^i\right|^2. 
    \label{loss_2}
\end{gather}
After fitting the IANN, the resulting approximation of $f$ can be visualized by plotting $\Hat{g}$ as a function of $x_j$ and $\Hat{h}$, as in Figure \ref{fig:harmonic_intro}. The fitted IANN serves as an approximation of $f$. To assess whether the  approximation is adequate, we generate a large random sample in the input space to serve as a test set and compute the test $r^2$ by comparing the IANN test predictions $\hat{f}$ with the actual function values $f$ at the test inputs.

\begin{figure}[htbp]
\centering
\includegraphics[width=\textwidth]{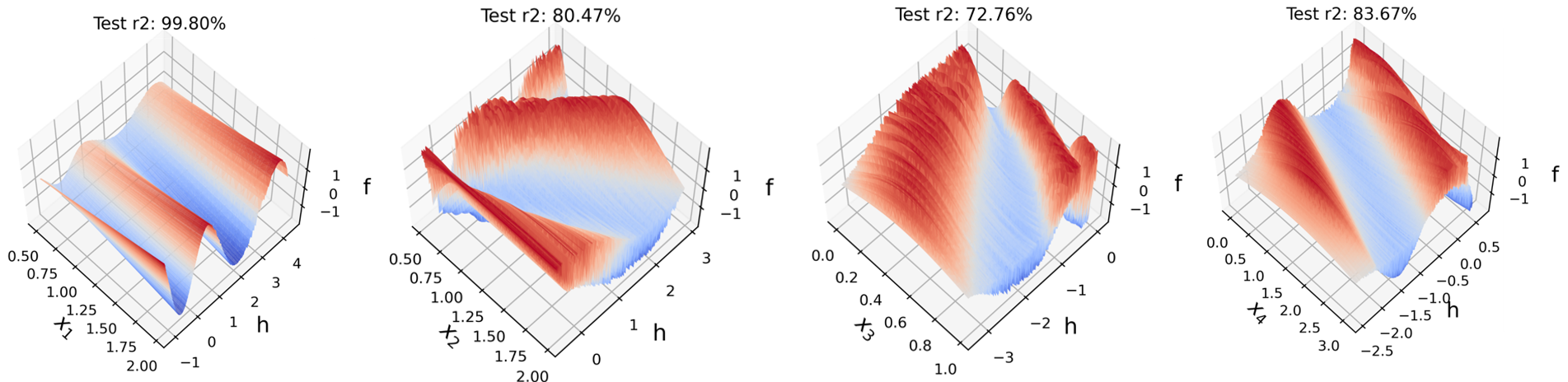}
\caption{Top-level (non-hierarchical) visualization plots of the harmonic wave function in Eq. (\ref{eqn:harmonic2}) with each input singled out in turn. The test $r^2$ is shown above each plot to evaluate the accuracy of IANN approximation represented by the plot.}
\label{fig:harmonic_first_level}
\end{figure}

Returning to the harmonic wave function defined in Eq. (\ref{eqn:harmonic2}), we (nonhierarchically) single out each of the four input variables one-by-one to serve as $x_j$ and fit the IANN architecture in Figure \ref{fig:exploration1}. The resulting four 3D plots (each of $g(x_j,h)$ as a function of $x_j$ and $h$ for the four different $x_j$) are displayed in Figure \ref{fig:harmonic_first_level}. The top left visualization plot has the highest test $r^2$ ($99.8\%$), which suggests that the harmonic wave function can be well approximated by the structure (\ref{function}) after singling out the amplitude variable $x_1$. This is consistent with the structure of $f$ in (\ref{eqn:harmonic2}) (which in practice would typically be a more complex and less directly interpretable function, e.g., a surrogate model fit to some complex computer simulation output), since (\ref{eqn:harmonic2}) can be written as $f(\boldsymbol{x}) = x_1* \sin \left[h(\xbs{1})\right]$ with   $h(\xbs{1}) = (2 \pi / x_2) *x_3+x_4$. From the top-left plot it is clear that the effect of $x_1$ is linear for any fixed $\xbs{1}$, although the effect of $x_1$ depends strongly on the value of $\xbs{1}$. For example, for some $h(\xbs{1})$ the effect of $x_1$ is positive (positive slope for the linear trend), and for other values the effect is negative. Moreover, the slope varies periodically as a sinusoidal function of $h(\xbs{1})$. Evidently, the IANN representation of $h(\xbs{1})$ captures the actual function $(2 \pi / x_2) *x_3+x_4$ quite well, which is more clear from the hierarchical IANN visualization presented later in Section \ref{sec:hierarchical_harmonic}.  The top-left plot in Figure \ref{fig:harmonic_first_level} also illustrates how the approach can be used to understand interactions between inputs, as $x_1$ and $h(\xbs{1})$ clearly have a strong interaction, by which the effect of $x_1$ changes from positive to negative and vice-versa as $h(\xbs{1})$ varies. 

For the other three plots, the relatively low test $r^2$ values suggest that singling out those variables in the IANN cannot approximate the structure (\ref{function}) well. Therefore, those plots are less reliable and should be used with caution when interpreting the effects of inputs. In this situation, we recommend using only the top-left plot of $g(x_1,h(\xbs{1}))$ (with test $r^2 = 99.8\%$) at the top level and then using the hierarchical decomposition in Section \ref{sec:hierarchical} to visualize the effects of $\{x_2, x_3, x_4\}$. We do this in Section \ref{sec:hierarchical_harmonic} in a continuation of this example. As pointed out after Theorem \ref{thm}, for a particular $x_j$ the latent function $h$ may be too complex to estimate via a neural network with limited number of layers and nodes (and also too complex to easily visualize in subsequent levels), which would be revealed by a low associated test $r^2$. But for all real examples that we have considered, the 3D visualization plot of $g(x_j, h(\xbs{j}))$ for at least one $x_j$ has sufficiently high test $r^2$ to render the visualization reliable. If multiple plots have high test $r^2$, users may rely on any or all of them to interpret the effects of inputs.

\subsection{A connection to PD and ICE plots}
\label{sec:ICE}

The most popular method for visualizing the effects of the input variables is partial dependence (PD) plots, which were first introduced in \citet{friedman2001greedy}. Later, \citet{goldstein2015peeking} proposed ICE plots, which enhance PD plots by displaying a collection of curves for each fixed combination of the omitted input variables. Here we compare and draw a connection between IANN, PD, and ICE plots using the harmonic wave function to illustrate.

\begin{figure}[htbp]
    \centering
    \includegraphics[width=\textwidth]{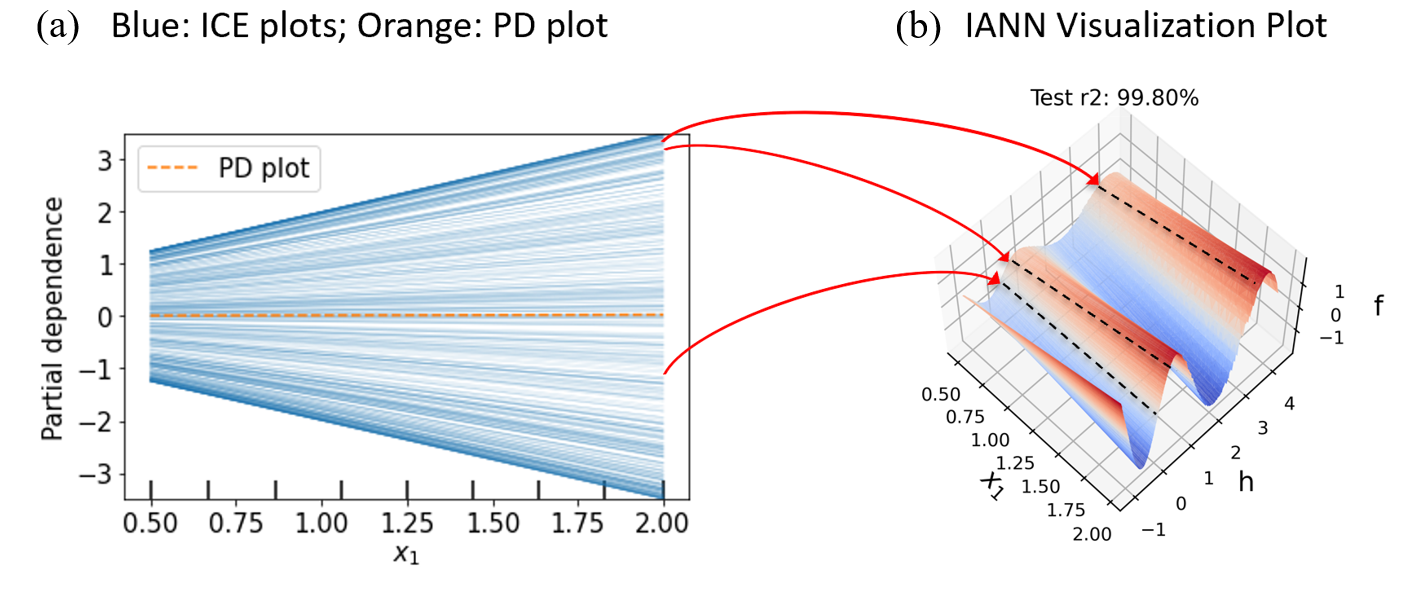}
\caption{Connection between (a) existing methods (ICE and PD plots) and (b) our IANN for visualizing the effects of the amplitude variable $x_1$ in the harmonic wave function. The IANN visualization can be viewed as smoothly piecing together the individual curves in the ICE plots to create a more interpretable 3D surface. The three dashed curves in (b) correspond to the three ICE plot curves indicated by connecting arrows in (a).}
\label{fig:harmonic_compare}
\end{figure}

Figure \ref{fig:harmonic_compare}(a) shows both the PD plot and the ICE plot for the variable $x_1$ using the open-source Python package, \textit{scikit-learn} \citep{scikit-learn, sklearn_api}. Notice that the PD plot, represented by the dashed line, is the average of all the curves in the ICE plots. There is no tight, direct connection between our IANN plots and ALE plots. Their only connection is via the connection between IANN plots and PD plots, and recognizing that ALE plots are intended to produce something similar to PD plots, albeit in a manner that is much more computationally efficient and less prone to problems when the inputs are highly correlated. Since the inputs have little correlation in the harmonic wave example, the ALE plot (omitted for brevity) is very similar to the PD plot in Figure \ref{fig:harmonic_compare}(a).  The ICE plots for this example are all straight lines with different slopes, which correctly suggests that the effect of $x_1$ is linear for fixed $\xbs{1}$ and that $x_1$ interacts with the other variables (since the slopes vary). In comparison, Figure \ref{fig:harmonic_compare}(b) shows the IANN plot of $g(x_1, h(\xbs{1}))$ for the same example. Note that each individual curve in the ICE plot represents $f$ as a function of $x_1$ for a fixed $\xbs{1}$. The IANN plot can be viewed as smoothly piecing together all of the ICE plot curves to create a 3D surface that serves as a more structured and interpretable visualization of the effect of $x_1$ on $f$ and of how $x_1$ interacts with some function ($h$) of the remaining variables $\xbs{1}$. Three individual curves for three fixed $h(\xbs{1})$ values are shown as dashed curves in Figure \ref{fig:harmonic_compare}(b), and the three corresponding ICE plot curves for the same three $\xbs{1}$ values are indicated in Figure \ref{fig:harmonic_compare}(a) by the connecting arrows. Compared to the 2D ICE plot visualization, the 3D IANN visualization provides a clearer picture of the function $f$ in this case.

\section{Hierarchical IANN for Visualizing the Joint Effects of All Inputs}
\label{sec:hierarchical}

In addition to being used as a stand-alone visualization of the effect of a selected input $x_j$, the IANN structure described in Section \ref{sec:first_level} can be used in a hierarchical manner to similarly decompose and visualize $h$, and so on. We present two versions of hierarchical decomposition, each of which is intended to visualize the joint effects of all $d$ inputs. In Section \ref{sec:ovh}, we introduce the original variable hierarchical (OVH) structure, and in Section \ref{sec:dash} we introduce the disjoint active subspace hierarchical (DASH) structure, which is in some sense a generalization of the OVH structure that can simplify visualization when the various functions involved are functions of certain linear combinations of the inputs.

\subsection{Original variable hierarchical (OVH) structure}
\label{sec:ovh}

In the structure (\ref{function}), $h$ is a function of all the inputs in $\xbs{j}$. To visualize how $h$ depends on $\xbs{j}$, we can decompose the functions hierarchically (adding subscripts to $g$ and $h$ to indicate the hierarchical level) via
\begin{align}
    f(\boldsymbol{x}) & \approx g_1(x_{j_1}, h_1(\boldsymbol{x}_{\setminus{j_1}})) \notag \\
    h_{i-1}(\boldsymbol{x}_{\setminus{(j_1, \cdots, j_{i-1})}}) & \approx g_{i}(x_{j_{i}}, h_{i}(\boldsymbol{x}_{\setminus{(j_1, \cdots, j_{i})}})),  \quad i = 2, \cdots, d-1,
    \label{eqn:f}
\end{align}
with $h_{d-1}(x_{j_d}) \defeq x_{j_d}$. Here, $\{j_1, j_2, \cdots, j_d\}$ represent a permutation ordering of the input indices $\{1, 2, \cdots, d\}$ that will be determined as a preprocessing step prior to fitting the hierarchical IANN. The algorithm for determining the ordering of the inputs is described in Section \ref{sec:algorithm}. Similar to the notation $\xbs{j}$, the notation $\boldsymbol{x}_{\setminus{(j_1, \cdots, j_{i})}}$ represents the input variables excluding $\{x_{j_1}, \cdots, x_{j_i}\}$. Each function $g_i$ is the approximation of the function $h_{i-1}$ with two inputs: $x_{j_{i}}$ and $h_{i}(\boldsymbol{x}_{\setminus{(j_1, \cdots, j_{i})}})$. Each function $h_i$ depends on the order $\{j_1, j_2, \cdots, j_{i}\}$ of the inputs selected in the prior iterations, although we omit this in the notation for simplicity.
Since the function $h$ in Theorem \ref{thm} is also continuous, we can repeatedly apply Theorem \ref{thm} to the function $h_{i-1}$ at each level to justify the hierarchical decomposition (\ref{eqn:f}).

We refer to the first equation in (\ref{eqn:f}) as the \emph{Level 1 representation} and the corresponding 3D plot or 2D heatmap ($f(\boldsymbol{x})$ as a function of $x_{j_1}$ and $h_1(\xbs{j_1})$) as the \emph{Level 1 plot}. Similarly, we refer to the $i^{th}$ equation in (\ref{eqn:f}) as the \emph{Level $i$ representation} and its corresponding 3D plot or 2D heatmap ($h_{i-1}$ as a function of $x_{j_i}$ and $h_i$) as the \emph{Level $i$ plot}, for $i = 2, \cdots, d-1$. This approach produces a total of $d-1$ hierarchical visualization plots that can collectively be used to understand the joint effects of all $d$ inputs. 

\begin{figure}[htbp]
\includegraphics[width=\textwidth]{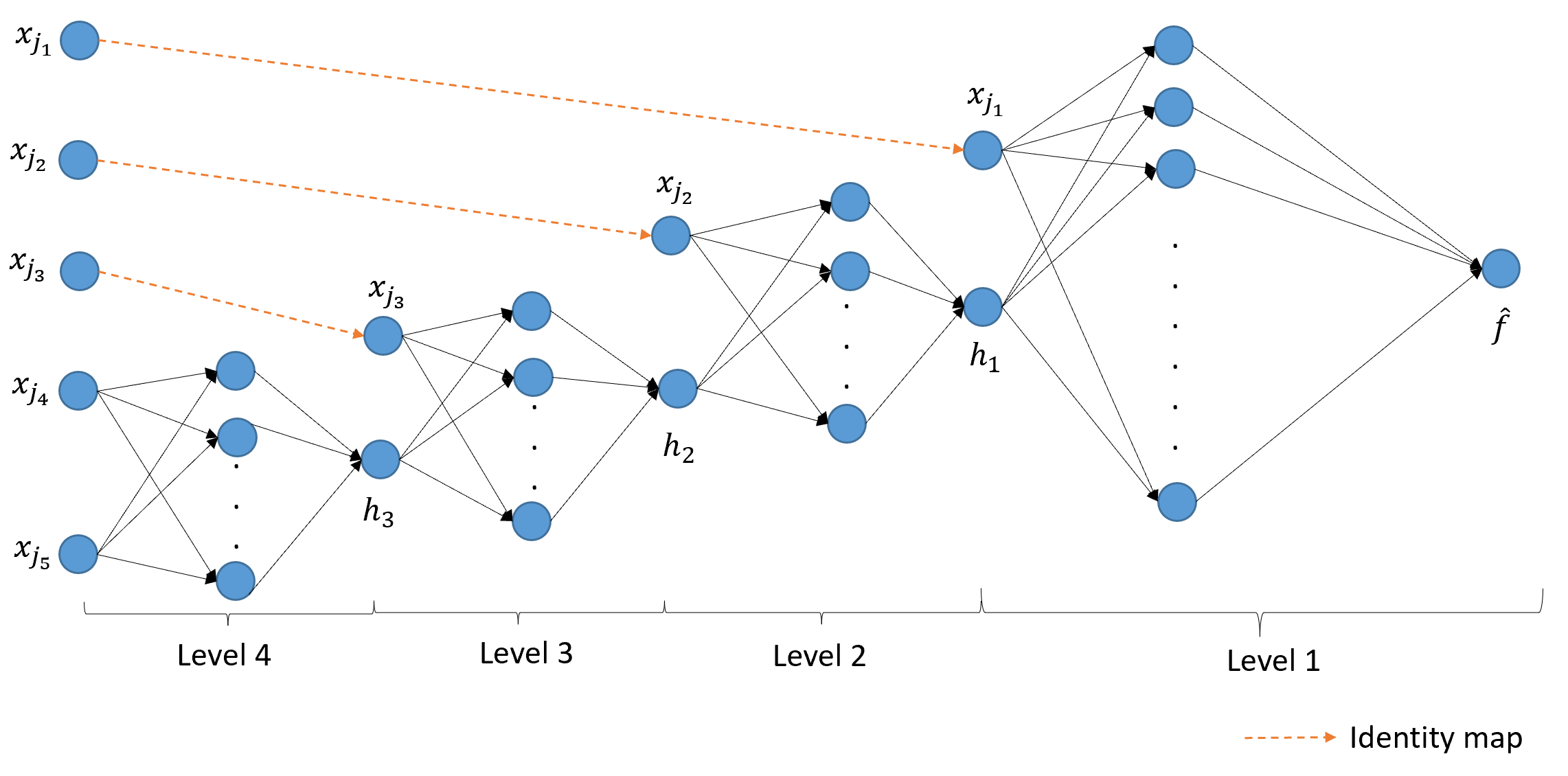}
\caption{IANN architecture for the OVH structure for the case of $d = 5$ input variables. }
\label{fig:general_architecture}
\end{figure}

To estimate all the $h_{i}$ functions simultaneously, we use an IANN architecture with multiple bottle-neck layers as illustrated in Figure \ref{fig:general_architecture} for the case of $d=5$ inputs. The dashed arrows represent the identity map. To simplify the notation, we omit the hat symbols on the $h_i$'s in the IANN architecture.
In Figure \ref{fig:general_architecture}, the IANN architecture consists of 5 input variables and 4 levels in total. The fourth level is just a fully connected neural network with two input variables and an output that represents the (estimated) function $h_3(x_{j_4}, x_{j_5})$. Likewise, the third level has bottleneck input layer with two inputs ($x_{j_3}$ and $h_3$) and $h_2$ as its output, the second level has bottleneck input layer with two inputs ($x_{j_2}$ and $h_2$) and $h_1$ as its output, and the first level has $x_{j_1}$ and $h_1$ as its two bottleneck inputs and the estimated/approximated $f$ as its output. 

Similar to the loss function defined in (\ref{loss_2}), we use $L_2$ loss to minimize the mean squared error between the IANN output $\Hat{f}$ and the original function $f$. By estimating the original function $f(\boldsymbol{x})$ and all the latent functions $h_i$ simultaneously, the error is less prone to accumulate through the different levels when $d$ is larger.

\begin{figure}[H]
\includegraphics[width=.8\textwidth]{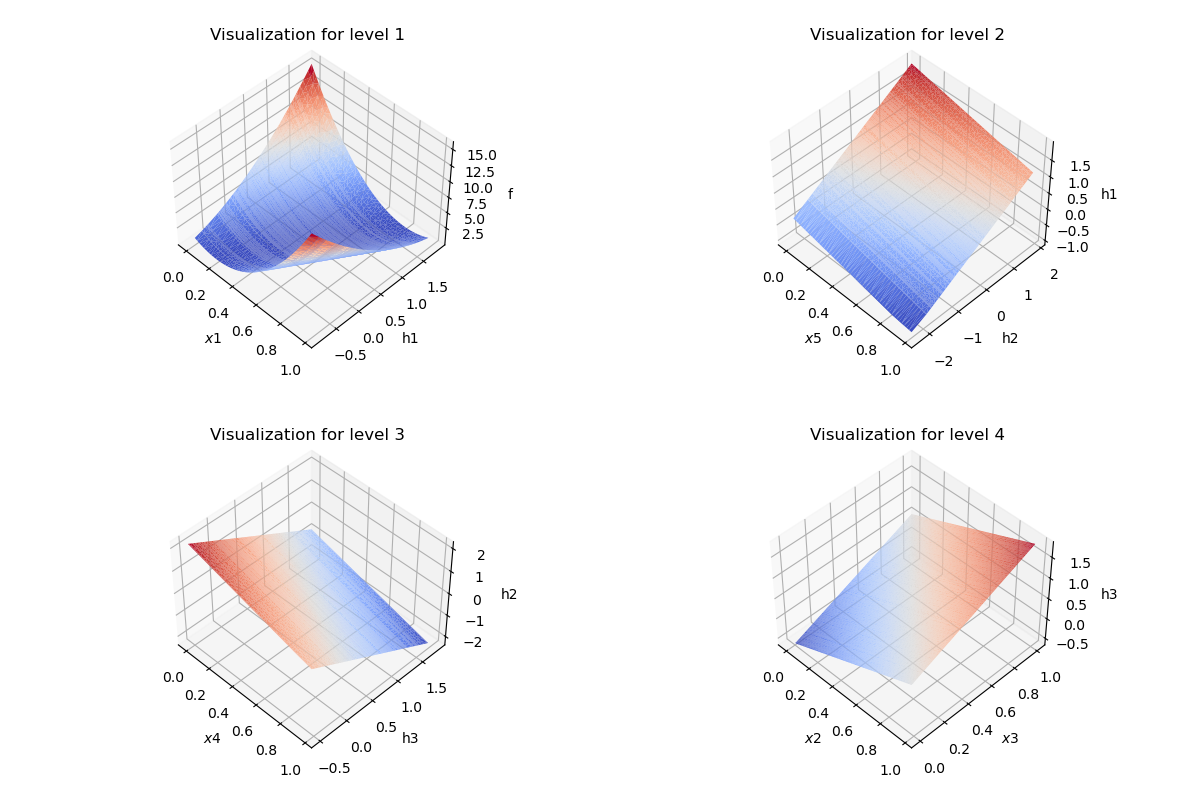}
\caption{IANN visualization plots for the example $f$ in Eq. (\ref{eqn:1}). The plots are to be read in numerical order of the levels, from the top left to the bottom right.}
\label{fig:general}
\end{figure}

To illustrate the OVH IANN approach, consider the following example:
\begin{equation}
    f(\boldsymbol{x}) = (5x_1 + x_2 +x_3 + x_4 + x_5-4.5)^2, \qquad \boldsymbol{x} \in [0, 1]^5.
    \label{eqn:1}    
\end{equation}
In this case, the input ordering $\{j_1, j_2, \cdots, j_5\}$ that satisfies the structure (\ref{eqn:f}) is not unique. In fact, any permutation of the input variables satisfies (\ref{eqn:f}). For example, if we choose $j_1 = 2$, the response can be represented as:
\begin{equation}
    f(\boldsymbol{x}) = g_1(x_2, h_1(\xbs{2})) =  (x_2 + h_1(\xbs{2}))^2,
    \label{eqn:gh}
\end{equation}
where $h_1(\xbs{2}) = 5x_1+x_3 + x_4 + x_5-4.5$. Following the same logic, $h_1$ can be decomposed similarly regardless of which remaining input is chosen as $x_{j_2}$, and so on. Note that the above arguments would hold if any of the five inputs were selected as $x_{j_1}$. In the following, we use the ordering $(j_1, j_2, j_3, j_4, j_5) = (1, 5, 4, 2, 3)$, which was determined by the algorithm described in Section \ref{sec:algorithm}. The visualization 3D plots are shown in Figure \ref{fig:general}. The test $r^2$ was $99.97\%$ indicating that the OVH structure (\ref{eqn:f}) provides a good approximation of $f$.

The following are salient points taken from the IANN visualization plots in Figure \ref{fig:general}. From the Level 1 plot the function $f(\boldsymbol{x})$ appears quadratic in $x_{1}$ when we hold $h_1$ fixed, which is consistent with the true function $f(\boldsymbol{x}) = (5x_{1} - h_1)^2$ if we take $h_1 =  -(x_2 +x_3 + x_4 + x_5-4.5)$. The functions $g_1$ and $h_1$ in (\ref{eqn:gh}) are not unique, of course, since we can incorporate an additive and/or multiplicative constant into $h_1$ and modify $g_1$ accordingly without changing the function $g_1(x_1, h_1(\xbs{1}))$. This, however, does not change the interpretation of the joint effects of the input variables. Further regarding the interpretation of the Level 1 plot, the function $f$ reaches its minimum at a value of $x_1$ that increases linearly as $h_1$ increases.

In order to understand the effects of the other inputs $\xbs{1}$ on $f$, we must discern two things from the plots in Figure \ref{fig:general}:  (i) The effect of $h_1$ on $f$, and (ii) the effect of the variables in $\xbs{1}$ on $h_1$. Regarding the former, from the Level 1 plot, for fixed $x_1$ the function $f$ appears roughly  quadratic in $h_1(\xbs{1})$. In order to understand the effects of $\xbs{1}$ on $h_1$, one must view the subsequent level plots. From the Level 2, 3, and 4 plots in Figure \ref{fig:general}, each function $h_i$ is approximately linear in its two arguments $x_{j_{i+1}}$ and $h_{i+1}$, which means that $h_1$ is linear in $\xbs{1}$. 

To further illustrate how the IANN level plots can help understand the behavior of $f$, supposed one wanted to select the values of $\boldsymbol{x}$ that maximize $f$. From the Level 1 plot, this occurs at the two corner points $(x_1, h_1) = (0, 1.5)$ or $(1, -1)$. If we focus on the former, from the Level 2 plot, $h_1 = 1.5$ occurs for $(x_5, h_2) = (0, 2)$. In turn, from the Level 3 plot, $h_2 = 2$ occurs for $(x_4, h_3) = (0, -0.5)$. Finally, from the Level 4 plot, $h_3 = -0.5$ occurs for $(x_2, x_3) = (0,0)$. Therefore, the maximum of $f$ at $(x_1, h_1) = (0, 1.5)$ corresponds to $x_1 = x_2 = \cdots = x_5 = 0$. Following the same procedure for the other maximum at $(x_1, h_1) = (1, -1)$, we find that this corresponds to $x_1 = x_2 = \cdots = x_5 = 1$. Both of these cases correspond to the true maxima of Eq.  (\ref{eqn:1}).

 Moreover, since $h_1(\xbs{1})$ is just a linear function of the remaining variables $x_2, x_3, x_4, x_5$ from the plots for Levels 2, 3, and 4, and since $f(\boldsymbol{x})$ is quadratic in $h_1(\xbs{1})$ and in $x_1$, we can conclude that $f$ is quadratic in each variable.

\subsection{Disjoint active subspace hierarchical (DASH) structure}
\label{sec:dash}

In this section, we propose an alternative hierarchical structure that can be viewed as a generalization of the OVH structure. The DASH structure assumes the hierarchical functions $h_i$'s are functions of certain disjoint linear combinations of the input variables:


\begin{align}
    f(\boldsymbol{x}) & \approx g_1(v_{1}, h_1(\boldsymbol{v}_{\setminus{1}})), \notag \\
    \label{eqn:g}
    h_{i-1}(\boldsymbol{v}_{\setminus{(1, \cdots, {i-1})}}) & \approx g_{i}(v_{{i}}, h_{i}(\boldsymbol{v}_{\setminus{(1, \cdots, {i})}})),  \qquad i = 2, \cdots, p-1, \\
    \mbox{where} \quad  v_{i} & = \boldsymbol{\beta}_i^T \boldsymbol{x}_{J_i}, \quad i = 1,2, \cdots, p \leq d. \notag
\end{align}
Here $\{J_1, \cdots, J_p\}$ are disjoint index sets, i.e., $J_i \bigcap J_j = \emptyset, \; \forall 1 \leq i \neq j \leq p$, and $\bigcup \limits_{i=1}^p J_i = \{1,2,\cdots, d\}$, and $\boldsymbol{x}_{J_i}$ denotes the input variables with indices in ${J_i}$. Note that the DASH structure (\ref{eqn:g}) reduces to the OVH structure when $p = d$.


Theorem \ref{thm} also applies to structure (\ref{eqn:g}) if we substitute the $x$'s with the $v$'s since the disjoint linear combinations of inputs (the $v$'s) take values in closed intervals. Similar to the proof for the OVH structure, one can repeatedly apply the theorem to $f$ and the $h_i$'s to show that any continuous function $f$ on $[0,1]^d$ can be arbitrarily closely approximated by the structure (\ref{eqn:g}) for some continuous functions $g_i$ and $h_i$. The algorithms we describe in Appendix A automatically find the number of disjoint linear combinations $p$, the $v$'s in (\ref{eqn:g}), and the order of the $p$ disjoint linear combinations $(v_{1}, v_{2}, \cdots, v_{p})$ that gives a good approximation of $f$ by (\ref{eqn:g}).

Similar to what was done for the OVH structure, we can make a series of 3D plots to visualize $f$ in (\ref{eqn:g}) as a function of the $v$'s, noting that the $v$'s are easy to understand as functions of the inputs, since they are disjoint linear combinations of the inputs. More specifically, in the Level 1 plot we visualize $f$ as a function of $v_{j_1}$ and $h_1$, and then further visualize $h_{i-1}$ as a function of $v_{j_{i}}$ and $h_{i}$ at the $i^{th}$ level for $i = 2,3, \cdots, p-1$. For many real examples that we have considered, Eq. (\ref{eqn:g}) with $p << d$ provides a close approximation of $f$. The advantage of this is that we reduce the number of hierarchical 3D visualization plots from $d- 1$ to $p- 1$, which substantially simplifies the interpretation when $p << d$.

\begin{figure}[htbp]
\includegraphics[width=\textwidth]{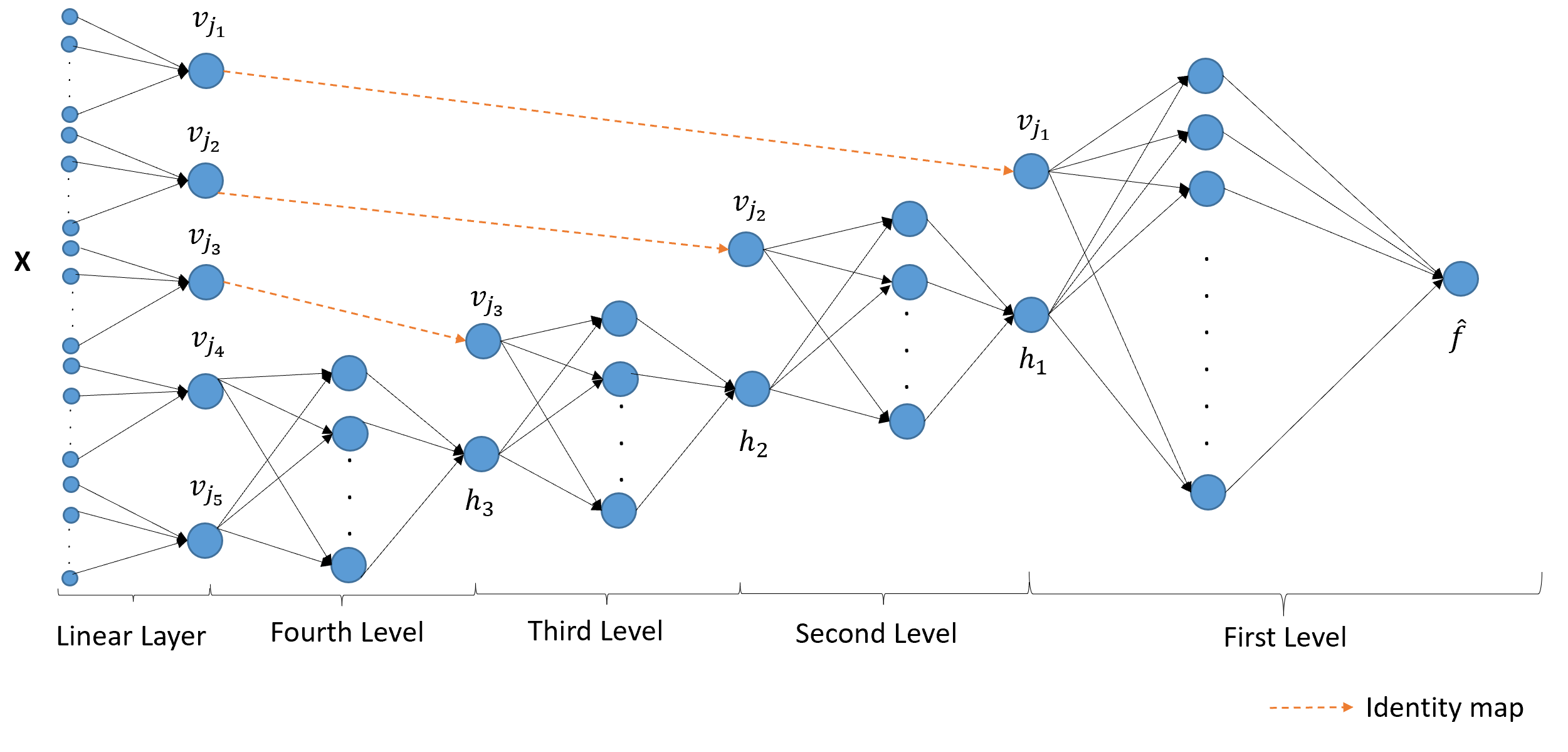}
\caption{Illustration of the IANN architecture for the DASH structure with $p = 5$.}
\label{fig:DASH_architecture2}
\end{figure}

To represent (\ref{eqn:g}) with an IANN architecture, we add one linear layer in front of the first layer in the OVH structure to learn the $p$ disjoint linear combinations $(v_1, v_2, \cdots, v_p)$ and the underlying disjoint input groups, as illustrated in Figure \ref{fig:DASH_architecture2} for the case of $p = 5$. Each linear combination $v_{i} = \boldsymbol{\beta}_i^T \boldsymbol{x}_{J_i}$ is represented by a single node in this layer with no bias and a linear activation function. The coefficients $\boldsymbol{\beta}_i$ in (\ref{eqn:g}) are the estimated weights in this linear layer. The remainder of the architecture is the same as in the IANN for the OVH structure shown in Figure \ref{fig:general_architecture}, except that the inputs to the subsequent layers are the disjoint linear combinations.  For visualization, we show the coefficients $\boldsymbol{\beta}_i$ below the 3D visualization plot of $h_{i-1}$ as a function of $v_{j_i}$ and $h_i$. 


\begin{figure}[htbp]
\includegraphics[width=\textwidth]{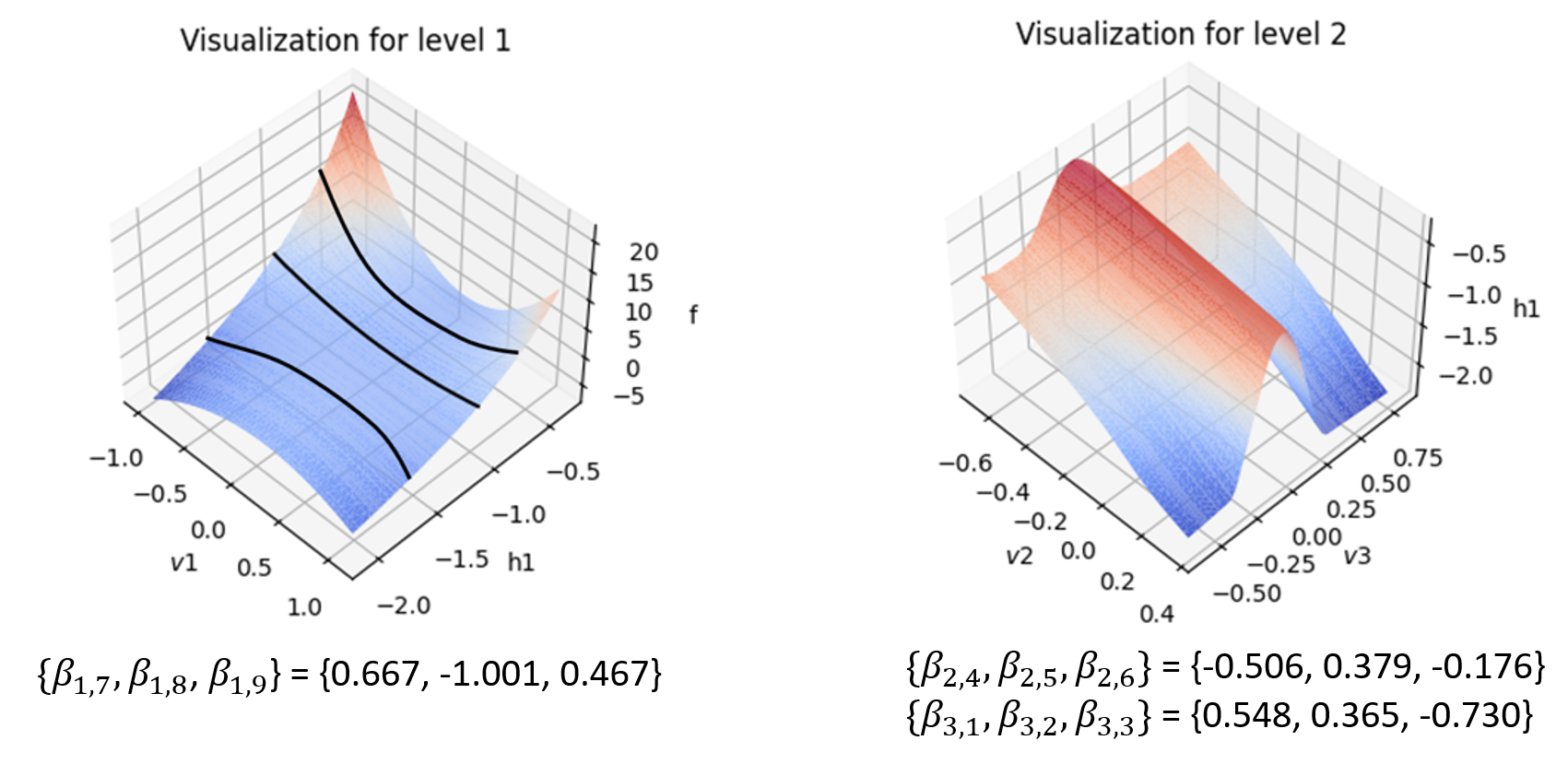}
\caption{DASH structure IANN visualization of $f$ in Eq. (\ref{eqn:DASH2}). The solid curves in the left figure are $f$ vs $v_1$ for fixed $h_1 = -1.5, -1.0$, and $-0.5$.}
\label{fig:DASH_eg2}
\end{figure}

To illustrate, consider the function:
\begin{equation}
    f(\boldsymbol{x}) = \left[7\exp{(-4(1.5 x_1 + x_2  - 2 x_3)^2)} + 2 x_4 -1.5 x_5  + 0.7 x_6 - 1.5\right](x_7 -1.5 x_8  + 0.7 x_9 - 0.3)^2,
    \label{eqn:DASH2}    
\end{equation}
for $\boldsymbol{x} \in [0,1]^9$, which can be represented as (\ref{eqn:g}) with $p = 3$ and

\begin{equation}
\begin{aligned}
   &    v_1  = x_7 -1.5 x_8  + 0.7 x_9, \\
   &    v_2  = 2 x_4 -1.5 x_5  + 0.7 x_6, \\
   &    v_3  = 1.5 x_1 + x_2  - 2 x_3, \quad  \mbox{and}\\
   &   f(\boldsymbol{x}) = (v_1 - 0.3)^2 * h_1(v_2, v_3),
\end{aligned}
\label{eqn:DASH2_coeff}    
\end{equation}
where $h_1(v_2, v_3) = \left[7\exp{(-4v_3^2)} + v_2 - 1.5\right]$.
We fit the DASH structure IANN to data from this function (the three disjoint linear combinations were automatically determined using the algorithm described in Appendix A and were not prespecified), the results of which are shown in Figure \ref{fig:DASH_eg2}. The test $r^2$ was $99.98\%$, indicating that the DASH structure (\ref{eqn:g}) provides a good approximation of $f$. 

Rather than using eight 3D plots to visualize and interpret the function $f(\boldsymbol{x})$ as with the OVH structure, we only need two 3D plots to visualize the effects of all the input variables with the DASH structure in Figure \ref{fig:DASH_eg2}. We show the coefficients $\beta_{i,j}$ of the inputs $x_j$ in each linear combination $v_i$ below each IANN visualization plot. From Figure \ref{fig:DASH_eg2}, we see that the estimated coefficients in each $\boldsymbol{\beta}_i$ have ratios that are similar to the true ratios defined in Eq. (\ref{eqn:DASH2_coeff}). For example, the true $v_1 = x_7 - 1.5x_8 + 0.7x_9$, and the estimated $v_1$ from fitting the IANN is $v_1 = 0.667x_7 - 1.001x_8 + 0.467x_9$, which is virtually identical except for a constant multiplicative factor.

Regarding interpreting the plots, first, from the top plot in Figure \ref{fig:DASH_eg2} the function $f(\boldsymbol{x})$ is non-monotonic (roughly quadratic) in $v_1$ when $h_1$ is fixed. Since $v_1$ is a linear combination of three input variables, $x_7, x_8, x_9$, we can conclude that each of these input variables have non-monotonic effects on $f(\boldsymbol{x})$, at least for some fixed values of the other two variables in this linear combination. From the same plot, when $v_1$ is fixed, function $f(\boldsymbol{x})$ monotonically increases in a nearly linear manner as $h_1$ increases. To visualize $h_1$, the Level 2 plot suggests that $h_1$ is monotonic and nearly linear in $v_2$ for each fixed $v_3$ but non-monotonic in $v_3$ for each fixed $v_2$. One conclusion from this is that $f$ is nearly linear in $x_4$, $x_5$, and $x_6$ (the inputs involved in $v_2$) with fixed $v_1$ but nonlinear in the other input variables.  Notice that the slight concave curvature in $f$ with respect to $h_1$ tends to cancel the slight convex curvature in $h_1$ with respect to $v_2$, so that $f$ is nearly linear in $v_2$ for $v_1$ fixed. 

Moreover, interactions between input variables can also be discerned from the plots. Understanding interactions (as opposed to additivity) between inputs is often important when interpreting models. We can tell whether an input variable in $v_1$ interacts with the remaining variables from the Level 1 IANN plot. Since the remaining variables reside either in $v_1$ or $h_1$, we consider two types of interactions: (i) interaction between variables in $v_1$ and in $h_1$, and (ii) interaction for variables within $v_1$. If neither interaction exists for that input variable, we can conclude that it has no interaction with all the remaining variables.

Regarding the interaction between $v_1$ and $h_1$, if the effect of $v_1$ on $f$ (e.g., the solid curves in the top left plot in Figure \ref{fig:DASH_eg2}) only changes by an additive constant as $h_1$ changes, then by definition, there is no interaction between $v_1$ and $h_1$ and thus no interaction between the inputs $\boldsymbol{x}_{J_1}$ and the inputs $\boldsymbol{x}_{\setminus J_1}$. Regarding the interactions within $v_1$, if $v_1$ has only linear effect on $f$ for each $h_1$ values, the input variables within $v_1$ do not interact with each other. Conversely, if $v_1$ has nonlinear effect on $f$ for at least some $h_1$ values, then there are interactions within $v_1$ based on the following argument. In this case, for the $h_1$ values for which the effect of $v_1$ is nonlinear, $\frac{\partial f(v_1, h_1)}{\partial v_1} = \tilde{g}(v_1, h_1)$ for some function $\tilde{g}$ that varies as $v_1$ varies, in which case the input variables $\boldsymbol{x}_{J_1}$ having nonzero $\boldsymbol{\beta}$ coefficients all interact with each other.  

From the IANN visualization in Figure \ref{fig:DASH_eg2}, we see from the three solid curves in the Level 1 plot that there is a strong interaction between $v_1$ and $h_1$. Since both $v_2$ and $v_3$ have an effect on $h_1$ from the second level plot, we conclude that the inputs in $\boldsymbol{x}_{J_1}$ interact with those in $\boldsymbol{x}_{J_2}$ and $\boldsymbol{x}_{J_3}$. Moreover, since the solid curves in the Level 1 plot are nonlinear, the inputs in $\boldsymbol{x}_{J_1}$ interact with each other.

As another example of how the IANN plots can be used, consider the so-called robust parameter design (RPD) problem \citep{taguchi1986introduction, robinson2004robust}, in which some of the input variables are controllable system/product design variables and others are noise variables that vary uncontrollably during system operation or product usage. The RPD goal is to find design values for the controllable inputs such that the function $f$ is least sensitive to variation in the uncontrollable inputs. To illustrate, suppose the inputs $\boldsymbol{x}_{J_1} = \{x_7, x_8, x_9\}$ comprising $v_1$ are noise variables, and the other inputs are controllable design variables. From the Level 1 plot in Figure \ref{fig:DASH_eg2}, we see that $f$ depends least strongly on $v_1$ when $h_1 \approx -1.0$. Consequently, if the inputs $\boldsymbol{x}_{\setminus{J_1}}$ on which $h_1$ depends are selected so that $h_1 \approx -1.0$ (e.g., by making use of the Level 2 plot in Figure \ref{fig:DASH_eg2}), then $f$ will be least sensitive to variation in the noise variables $\boldsymbol{x}_{J_1}$.

\section{IANN Algorithm Details}
\label{sec:algorithm}
This section describes how to determine the order of the input variables in the hierarchical IANN decomposition (i.e., which input variable appears at each level) and other details of the algorithm. Here, we mainly present algorithms for this purpose for the OVH IANN structures. The algorithms for the DASH IANN structures, the details of which can be found in Appendix A, are similar.


To best approximate $f$ with the OVH structure and make the visualization plots more reliable for interpretation, we determine the order of input variables sequentially by selecting the variable that results in the best approximation accuracy at each level in the OVH structure (\ref{eqn:f}). To see the intuition behind our approach for this, notice that if $f(\boldsymbol{x}) = g_1(x_{j_1}, h_1(\xbs{j_1}))$ in the first level of (\ref{eqn:f}), then by the chain rule,

\begin{equation}
    \frac{\partial f}{\partial \xbs{j_1}}(\boldsymbol{x}) = \frac{\partial g_1}{\partial \xbs{j_1}}(\boldsymbol{x}) = \frac{\partial g_1}{\partial h_1}(\boldsymbol{x}) \nabla h_1(\xbs{j_1}),
    \label{eqn:deri_1}
\end{equation}
where $\nabla$ denotes the gradient of a function with respect to its input arguments. Since $\frac{\partial g_1}{\partial h_1}(\boldsymbol{x})$ is a scalar, if we keep $\xbs{j_1}$ fixed and consider $\frac{\partial f}{\partial \xbs{j_1}}(\boldsymbol{x}) \in \mathbb{R}^{d-1}$ for many different $x_{j_1}$ values, they are all approximately colinear and differ only in their magnitude and sign. Therefore, as candidates for $x_{j_1}$, we consider the inputs whose gradients are the most colinear using principal component analysis (PCA) to measure the extent of colinearity. 

We note that although the functions $g$ and $h$ from the IANN structure appear in Eq. (\ref{eqn:deri_1}) (and in Eq. (\ref{eqn:deri_2}) below), they are not used in the algorithm for determining the ordering of the inputs in the IANN, which is a pre-processing step that uses the gradient of $f$ directly and requires no IANN fitting. Eqs. (\ref{eqn:deri_1}) and (\ref{eqn:deri_2}) are only used to justify the rationale behind our input ordering approach. We also note that finding the ordering of inputs $\boldsymbol{x}$ requires the function $f$ to be differentiable since we need to compute the gradient of $f(\boldsymbol{x})$ with respect to $\boldsymbol{x}$.

The following describes the details of this gradient projection algorithm for Level 1.
\noindent \textbf{Gradient projection algorithm:} (repeat steps 1-3 for each input variable $x_j, \; j = 1, 2, \cdots, d$)
\begin{enumerate}
    
    \item For input variable $x_j$, draw $N_{\setminus j}$ latin hypercube design (LHD) samples in the $\xbs{j}$ space, denoted by $\{\boldsymbol{x}_{\setminus j}^l; \; l = 1,\cdots, N_{\setminus j}\}$. Then draw $N_j$ evenly spaced points spanning the range of $x_j$ in the $x_j$ space, denoted by $\{x_j^m; \; m = 1,\cdots N_j\}$.
    
    \item For $l = 1, 2, \cdots, N_{\setminus j}$, calculate the gradient vectors $\boldsymbol{a}_{ j}^{m, l} = \nabla_{\xbs{j}} f(x_j^m, \boldsymbol{x}_{\setminus j}^l) \in \mathbb{R}^{d-1}, \; m = 1, 2, \cdots, N_j$, and stack them into the $N_j \times {(d-1)}$ matrix:
    \begin{equation}
        A^l_j = \begin{bmatrix}
            \left(\boldsymbol{a}_j^{1,l}\right)^T \\
            \vdots \\
            \left(\boldsymbol{a}_j^{N_j,l}\right)^T
        \end{bmatrix},  \quad \mbox{for }  l = 1, 2, \cdots, N_{\setminus j}.
        \label{eqn:gpa_level1_1}
    \end{equation}
    
    \item Use PCA to find the eigenvector ($\boldsymbol{z}^l_j$) corresponding to the largest eigenvalue of $\left(A^{l}_j\right)^T A_j^l$ for each $l = 1, 2, \cdots, N_{\setminus j}$, and normalize the eigenvector such that $\left|\left|\boldsymbol{z}^l_j\right|\right| = 1$. Compute the error between the gradient vectors $\boldsymbol{a}_j^{m,l}$ and their projections onto $\boldsymbol{z}^l_j$, defined as
    \begin{equation}
        E^l_j = \sum \limits_{m = 1}^{N_j} \left|\left| \boldsymbol{a}_j^{m,l} - \left[\boldsymbol{z}^l_j \left(\boldsymbol{z}^l_j\right)^T \right] \boldsymbol{a}_j^{m,l} \right|\right|^2, \quad \mbox{for }  l = 1, 2, \cdots, N_{\setminus j}.
        \label{eqn:gpa_level1_2}
    \end{equation}
    Then, compute the gradient projection error ($E_j$) for $x_j$ via
    \begin{equation}
        E_j = \frac{1}{N_{\setminus j}} \sum \limits_{l=1}^{N_{\setminus j}} \frac{E^l_j}{\Omega_j^l},
        \label{eqn:gpa_level1_3}
    \end{equation}
    where     
    \begin{equation}
        \Omega_j^l = \frac{1}{N_{ j}} \sum \limits_{m=1}^{N_j} \left|\left|\boldsymbol{a}_j^{m,l}\right|\right|^2  \qquad \mbox{for } l = 1, 2, \cdots, N_{\setminus j}.
        \label{eqn:gpa_level1_4}
    \end{equation}
    Note that we use the normalization factor $\Omega_j^l$ in (\ref{eqn:gpa_level1_3}) to express the projection error $E_j^l$ relative to the average squared length of the gradient vectors for that $l$.

    \item Finally, we choose the input variable $x_{j_1}$ as the one having the smallest projection error, i.e., $j_1 = \mathop{\arg\min}\limits_{j} \{E_j: j = 1, 2, \cdots, d\}$.
\end{enumerate}

We use the maximin criterion \citep{johnson1990minimax,mckay2000comparison} to draw the LHD samples throughout the paper, which aims to maximize the minimum distance between any two samples in the LHD to enhance the space-filling property. Also note that the entire IANN modeling procedure uses LHDs for two different purposes: First, in the gradient projection algorithm above for determining the input ordering, and then again to generate the training data for the IANN model fitting (see Appendix B). For the former, we have found that a standard LHD works fine. For the latter, we have found that the modified LHD in Appendix B consistently works better than a standard LHD.

The algorithm for selecting the input for each subsequent level is similar: Suppose we have determined the order of input variables in the previous levels, $\{x_{j_1}, \cdots, x_{j_{i-1}}\}$ for some $i > 1$. Similar to the procedure for finding $x_{j_1}$, for each candidate input index $j_i \in \{1, 2, \cdots, d\} \setminus \{j_1, \cdots, j_{i-1}\}$, we consider the gradient of $f(\boldsymbol{x})$ with respect to the remaining inputs $\boldsymbol{x}_{\setminus{(j_1, \cdots, j_i)}}$, which by the chain rule is (if structure (\ref{eqn:f}) holds with an equality),

\begin{equation}
    \frac{\partial f}{\partial \boldsymbol{x}_{\setminus{(j_1, \cdots, j_i)}}}(\boldsymbol{x}) = \frac{\partial g_1}{\partial h_1}(\boldsymbol{x}) \frac{\partial h_1}{\partial h_2}(\xbs{j_1}) \cdots \frac{\partial h_{i-1}}{\partial h_i}(\boldsymbol{x}_{\setminus{(j_1, \cdots, j_{i-1})}}) \nabla h_i(\boldsymbol{x}_{\setminus{(j_1, \cdots, j_i)}}).
    \label{eqn:deri_2}
\end{equation}
Then, if we fix $\boldsymbol{x}_{\setminus{(j_1, \cdots, j_i)}}$ and vary $(x_{j_1}, \cdots, x_{j_i})$ by taking some LHD samples in the $\boldsymbol{x}_{\left(j_1, \cdots, j_i\right)}$ space, the vector $\nabla h_i(\boldsymbol{x}_{\setminus{(j_1, \cdots, j_i)}})$ in (\ref{eqn:deri_2}) is a constant vector, so that the gradient vectors $\frac{\partial f}{\partial \boldsymbol{x}_{\setminus{(j_1, \cdots, j_i)}}}(\boldsymbol{x})$ are all colinear and differ only in their magnitude and sign. Consequently, we apply the same gradient projection algorithm described above to the subsequent levels by substituting $\xbs{j_1}$ with $\boldsymbol{x}_{\setminus{(j_1, \cdots, j_i)}}$ and selecting $x_{j_i}$ to be the remaining input whose gradient vectors in Eq. (\ref{eqn:deri_2}) are the most colinear according to the gradient projection error measure analogous to (\ref{eqn:gpa_level1_3}). To avoid generating LHD samples repeatedly in each level, we only take LHD samples once in the $d$-dimensional input space and then project them into the lower dimensional spaces where needed.

\section{Numerical Experiments}
\label{sec:numerical}

This section provides two examples of visualizing functions for which we have a closed-form “ground truth” expression to compare to our IANN visualizations. The Supplementary Materials section provides an example in which the function is an actual surrogate model of a complex numerical simulation of the potential energy of a strain-loaded material sample.

\subsection{IANN for the borehole function}

To illustrate how IANN plots help with the visualization and interpretation of function, we use the borehole function
\begin{equation}
    f(\boldsymbol{x}) = 2 \pi x_1 \left(x_4 - x_6\right)\left[\log\left(\frac{x_2}{x_3}\right)\left(1 + 2\frac{x_7 x_1}{\log(\frac{x_2}{x_3}) x_3^2 x_8} + \frac{x_1}{x_5}\right)\right]^{-1},
    \label{eqn:borehole}
\end{equation}
which models the water flow between two aquifers. 
Although the functional form in (\ref{eqn:borehole}) is known, it is commonly treated as a black-box function for evaluating surrogate modeling methods \citep{an2001quasi, harper1983sensitivity, morris1993bayesian, simulationlib}. The response $f$ is the water flow rate between the aquifers. Here, $d=8$, $x_1 \in [0.05, 0.15]$ and $x_2 \in [100, \num{50000}]$ represent the radius of a borehole and its influence respectively, $x_3 \in [\num{63070}, \num{115600}] $ and $x_5 \in [63.1, 116]$ represent transmissivity of the upper and lower aquifer, $x_4 \in  [990, \num{1110}]$ and $x_6 \in [700, 820]$ represent the potentiometric head of the upper and lower aquifer, $x_7 \in [\num{1120}, \num{1680}]$ represents the length of the borehole, and $x_8 \in [\num{9855},\num{12045}]$ represents the hydraulic conductivity of the borehole.


Since the borehole function has eight inputs, each with different scales, we first use the min-max normalization to rescale the range of inputs to $[0,1]$ when fitting the IANN model. The OVH structure will hierarchically generate seven 3D visualization plots. Instead, we use the DASH structure to interpret the effects of all the input variables through four linear combinations and their hierarchical ordering, which were automatically determined using the algorithm described in Appendix A, which is similar to the one in Section \ref{sec:algorithm}.

Regarding computational expense, the algorithm took 96 seconds to determine the input ordering for the borehole example. The main computational expense is in fitting the single IANN after determining the input ordering, which took 245 seconds for the borehole example. The computational expense of fitting the single IANN is comparable to that of fitting standard neural networks with comparable numbers of parameters. One can apply cross-validation to choose the optimal hyperparameters in our IANN model, the computational expense of which is proportional to the time for fitting a single IANN.

\begin{figure}[htbp]
\centering
\includegraphics[width=\textwidth]{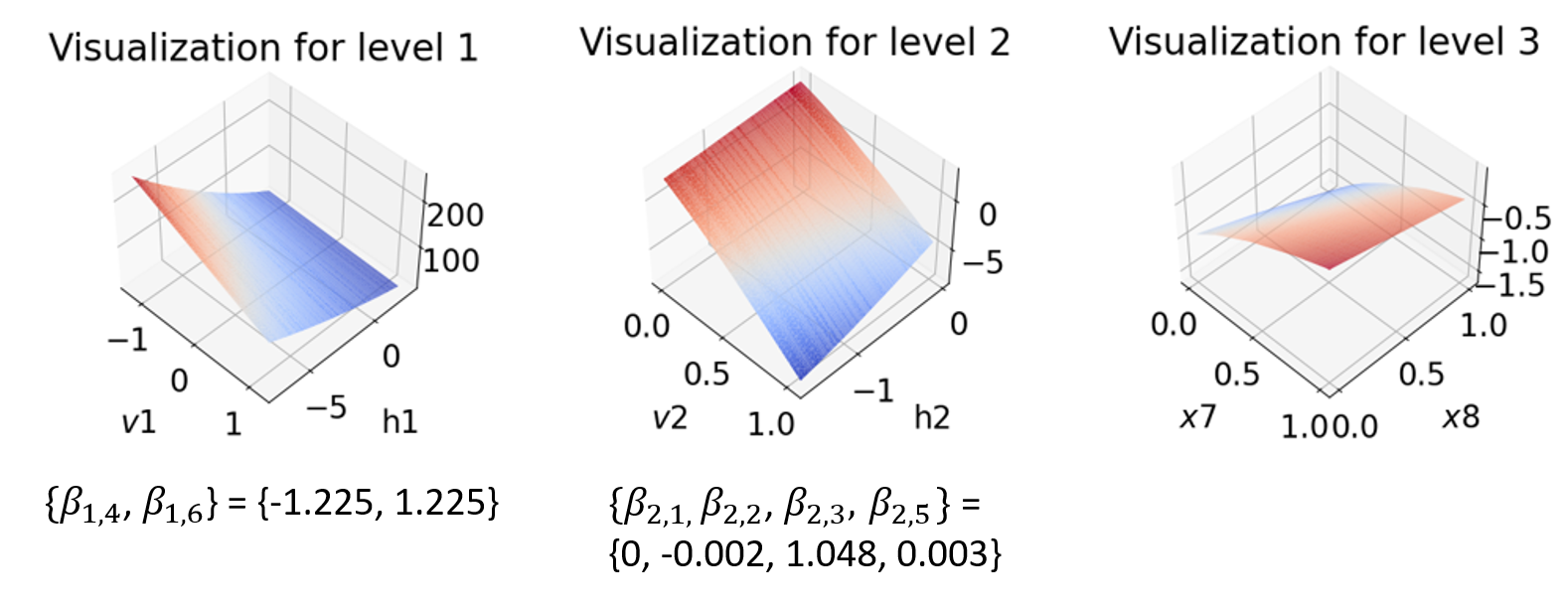}
\caption{DASH structure IANN visualization for the borehole function $f$ in Eq. (\ref{eqn:borehole}). The axes of the plots and the $\boldsymbol{\beta}$ coefficients listed below the plots correspond to the normalized inputs.}
\label{fig:borehole}
\end{figure}

In the event that users want to understand and interpret the effect of one input in particular, they can manually choose the ordering of the linear combination groups so that the first group contains the input variable of interest. Table \ref{tab:borehole} shows the ordered groups produced by the algorithms for the DASH IANN structure with the constraint that the first group is the one containing the input of particular interest. The eight rows show the resulting orderings with each of the eight inputs specified as being of particular interest. This way, one can more directly visualize its effect on $f$ from the Level 1 plot. The order of the remaining linear combinations is shown in the second column in Table \ref{tab:borehole}. For each of the orderings in Table \ref{tab:borehole}, we also show the resulting test $r^2$. Figure \ref{fig:borehole} shows the IANN visualization plot with the highest test $r^2$.


For comparison, we use global sensitivity analysis \citep{Herman2017} to compute the total sensitivity indices for $\{x_1, x_2, \cdots, x_8\}$, which are $\{0.0, 0.0, 0.174, 0.26, 0.0, 0.261,\\ 0.258, 0.063\}$. This suggests that $x_1, x_2$ and $x_5$ have little effect on the borehole function, which is also reflected in the IANN plot in Figure \ref{fig:borehole}.


\begin{table}[htbp]
    \centering
    \begin{tabular}{|c|c|c|}
        \hline
        index j & input groups and ordering & test $r^2$\\
        \hline
        1 & [[1, 2, 3, 5], [4, 6], [7], [8]] & 99.97\% \\
         \hline
        2 & [[1, 2, 3, 5], [4, 6], [7], [8]] & 99.97\%  \\
         \hline
        3 & [[1, 2, 3, 5], [4, 6], [7], [8]] & 99.97\%   \\
         \hline
        4 &  [[4, 6],[[1, 2, 3, 5],  [7], [8]]& 99.99\%  \\
         \hline
        5 & [[1, 2, 3, 5], [4, 6], [7], [8]] & 99.97\%  \\
         \hline
        6 & [[4, 6],[[1, 2, 3, 5],  [7], [8]] & 99.99\% \\
         \hline
        7 & [[7],[4, 6],[[1, 2, 3, 5],   [8]] & 99.94\%  \\
         \hline
        8 &  [[8],[4, 6],[[1, 2, 3, 5],   [7]] & 99.99\%   \\
         \hline
    \end{tabular}
    \caption{For the borehole function with the DASH structure assumed, a listing of the $p = 4$ disjoint linear combination groups and their orderings, which were automatically determined by algorithms for the DASH IANN structure under the constraint that the first group contains the specified input variable of particular interest, which is shown in Column 1.}
    \label{tab:borehole}
\end{table}

We used $\numprint{34000}$ training samples for fitting our IANN and $10^6$ randomly generated test input samples to compute the test $r^2$. The test $r^2 = 99.99\%$ for the Borehole function. The coefficients of the (normalized) inputs in each linear combination are listed below the plots in Figure \ref{fig:borehole}. Since the inputs have been normalized to $[0,1]$, comparing the magnitudes of the coefficients suggests the relative importance of the inputs involved in each linear combination. For example, from the Level 1 plot in Figure \ref{fig:borehole}, $v_1$ consists of two input variables, $x_4$ and $x_6$, and the corresponding coefficients have the same magnitude with different signs. This agrees with the formula for the borehole function in Eq. (\ref{eqn:borehole}), which can be written as $f(\boldsymbol{x}) = (x_4 - x_6) h_1(\xbs{(4,6)})$ for $h_1 = 2 \pi x_1\left[\log\left(\frac{x_2}{x_3}\right)\left(1 + 2\frac{x_7 x_1}{\log(\frac{x_2}{x_3}) x_3^2 x_8} + \frac{x_1}{x_5}\right)\right]^{-1}.$  From the Level 2 plot, the coefficients are almost zero except for the input variable $x_3$. This also agrees with the global sensitivity analysis, which shows that $x_1, x_2, x_5$ have little impact on the response $f$.

As a reference for comparison, the test $r^2$ for a linear model fitted to the borehole function is $94.68\%$, which suggests that the borehole function is almost linear over the specified input domain. The examples in the subsequent sections demonstrate the interpretability of the IANN visualization for functions having a higher level of nonlinearity. 

\subsection{IANN for the harmonic wave function}
\label{sec:hierarchical_harmonic}

Reconsider the harmonic wave function $f(\boldsymbol{x})$ in Eq. (\ref{eqn:harmonic2}), for which the IANN visualizations for only Level 1 were shown in Figures \ref{fig:harmonic_intro} and \ref{fig:harmonic_first_level}. We first considered the DASH structure to approximate $f$, but it found no relevant disjoint linear combinations other than the trivial one with $p = d$ and each $v_i$ a single input variable. Consequently, we use the OVH structure. More generally, we recommend that users first try the DASH structure, and if it does not produce any relevant linear combinations that result in an acceptably high test $r^2$, then users should use the OVH structure.

\begin{figure}[htbp]
\centering
\includegraphics[width=\textwidth]{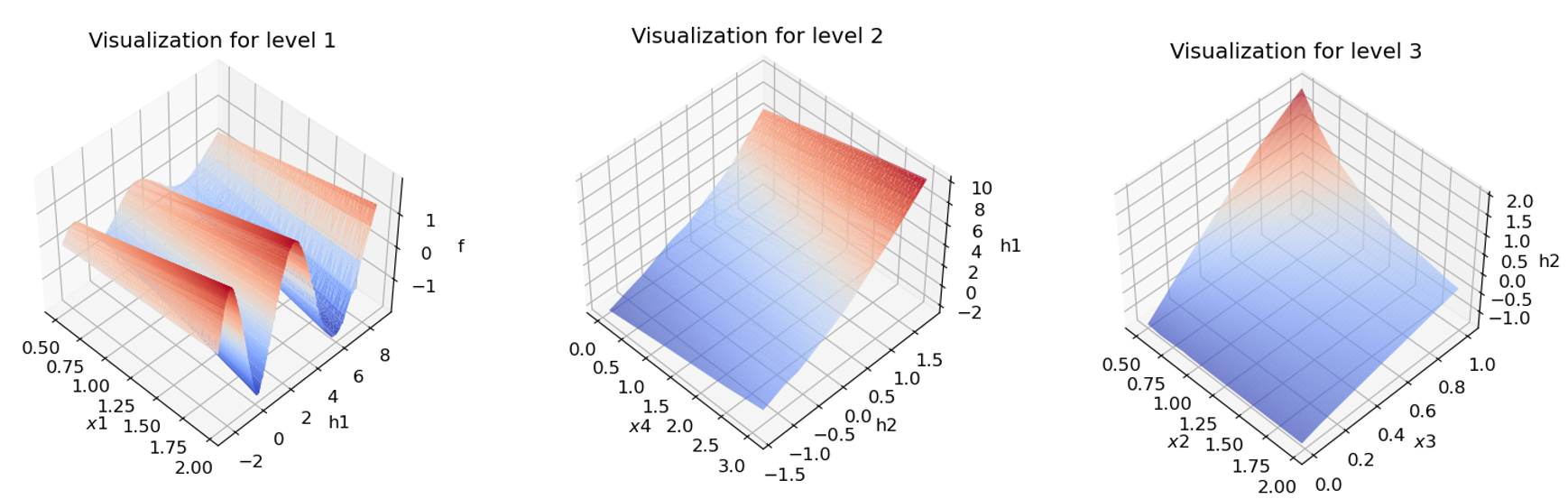}
\caption{OVH structure IANN visualization of harmonic wave function in Eq. (\ref{eqn:harmonic2}). The test $r^2 = 99.95\%$.}
\label{fig:harmonic}
\end{figure}

From the Level 1 plot in Figure \ref{fig:harmonic} (which is similar to Figure \ref{fig:harmonic_intro}), we see that $x_1$ has linear effect on $f(\boldsymbol{x})$ when the other variables are fixed, and $f(\boldsymbol{x})$ is sinusoidal for each fixed value of $x_1$, which agrees with Eq. (\ref{eqn:harmonic2}).  From the Level 2 plot,  $h_1(\xbs{1})$ is linear in $x_4$ and $h_2$, and $x_4$ and $h_2$ have little interaction, which suggests the additive structure $h_1(\xbs{1}) = \alpha_1 h_2(\xbs{(1,4)}) + \alpha_2 x_4$ for some constants $\alpha_1$ and $\alpha_2$. Therefore, $x_4$ reflects the phase angle in the harmonic function.  From the Level 3 plot, we can see that $h_2$ is linear in $x_3$ for each fixed $x_2$ and nonlinear in $x_2$ for some fixed $x_3$, especially when $x_3$ is higher. Moreover, it is clear 
from the Level 3 plot that $x_2$ has zero effect on $h_2$ when $x_3 = 0$, versus a large and nonlinear effect on $h_2$ when $x_3 = 1$, which also agrees with the ``true" expression for $h_2$ ($= 2 \pi\frac{ x_3}{x_2}$ up to some constants) from Eq. (\ref{eqn:harmonic2}).  Likewise from the Level 3 plot, as we decrease $x_2$, $x_3$ has a far larger effect on $h_2$, and thus a far larger effect on $h_1$ (from the Level 2 plot) and on $f$ (from the Level 1 plot). From this, for smaller $x_2$, $f$ undergoes more sinusoidal cycles as $x_3$ varies over its full range, which agrees with the observation from Eq. (\ref{eqn:harmonic2}) that $x_2$ represents the wavelength of the sinusoidal function. In this sense, the IANN plot truly reflects the physical interpretation of the input variables.

One can find more examples in the Appendix C that use our IANN approach to visualize actual black box functions that represent a surrogate model fit to the output of a black-box computer simulation of a physical system.

\section{Conclusions and Potential Extensions}
\label{sec:conclusion}
This paper introduced the IANN approach to visualize and interpret black box functions. We have shown that, theoretically, any continuous function on $[0,1]^d$ can be arbitrarily closely approximated by our more interpretable structure, which can be conveniently represented with the proposed IANN architecture. To visualize the effects of all the input variables, we developed two hierarchical structures (OVH and DASH), each of which can be represented with a particular IANN architecture. We have also developed algorithms to automatically determine the ordering of input variables with the goal of providing the best approximation to the original function for each hierarchical structure. We have used a number of examples to demonstrate the interpretability advantages of our IANN method. 

We envision several potential extensions of our IANN approach to either enhance the interpretation of the original function $f$ and/or to expand the class of $f$ to which our IANN provides a good approximation. One way to enhance interpretability with the IANN visualization plots is to develop customized graphical user interfaces (GUIs), with which users can change the value of each input variable $\{x_1, x_2, \cdots, x_d\}$ via slide-bar controls and interactively visualize the corresponding points $\{f, h_1, \cdots, h_{d-2}\}$ on the IANN visualization surfaces plotted in each level. Moreover, to facilitate the creation of 3D trellis plots of $f$ vs $\{x_j, x_l\}$ for some collection of fixed values of  $\boldsymbol{x}_{\backslash{(j,l)}}$ (which normally requires careful selection of many fixed $\boldsymbol{x}_{\setminus{(j,l)}}$ values when $d$ is larger), one could modify the IANN architecture to fit a model of the form
\begin{equation}
    f(\boldsymbol{x}) = g(x_j, x_l, h(\boldsymbol{x}_{\backslash{(j,l)}})),
    \label{eqn:extension1}
\end{equation} in which case users need only to select a few fixed values of the \textit{scalar} $h(\boldsymbol{x}_{\backslash{(j,l)}})$, instead of selecting many fixed values of the higher-dimensional $\boldsymbol{x}_{\setminus{(j,l)}}$. 

One might also consider removing the disjoint restriction on the linear combinations in the DASH structure, which would allow each input to appear in multiple combinations. The main reason we did not pursue this is that we are prioritizing interpretability of the effects of individual inputs over generality of the IANN approach. Having overlapping linear combinations would mean that the same input variable is present in multiple linear combinations, which makes the interpretation of that individual input variable less clear. Considering that the disjoint set restriction in the DASH structure becomes less restrictive and more general as the number of linear combinations increases (in the limiting case, each linear combination is a single input, in which case the DASH structure reduces to the OVH structure), we thought the disjoint restriction constitutes a reasonable tradeoff between generality and interpretability. As future work, we plan to explore this issue further, but as yet we are unsure how to preserve interpretability with overlapping linear combinations. 

A related extension is to use the modified architecture (\ref{eqn:extension1}) and simply plot $f$ vs $\{x_j, x_l\}$ with $h(\boldsymbol{x}_{\backslash{(j,l)}})$ controlled by a slide bar, which would be interpreted similarly to 3D trellis plots but completely avoids having to select any fixed values related to the omitted variables $\boldsymbol{x}_{\setminus{(j,l)}}$. Analogous to the OVH structure in Section \ref{sec:ovh}, one could further decompose $h$ in Eq. (\ref{eqn:extension1}) as a function of two additional inputs and a second latent function that is likewise controlled by a slide-bar in a Level 2 plot of $h$ versus the two additional inputs, and so on in subsequent levels. This extension would reduce the number of required levels by roughly half in the IANN visualizations and would also expand the applicability of the approach from functions that can be represented by (\ref{function}) to functions that can be represented by (\ref{eqn:extension1}). 

Another potential extension is to use a dichotomous tree IANN structure, for which the Level 1 plot is of $f(\boldsymbol{x}) = g_1(h_{1,1}(\boldsymbol{x}_{J_{1,1}}), h_{1,2}(\boldsymbol{x}_{J_{1,2}}))$ vs $h_{1,1}(\boldsymbol{x}_{J_{1,1}})$ and $h_{1,2}(\boldsymbol{x}_{J_{1,2}})$, where $J_{1,1}$ and $J_{1,2}$ are a disjoint partition of $\{1, 2, \cdots, d\}$. Each latent $h$ function can then be decomposed similarly in the subsequent levels to produce a tree-like nested set of 3D plots. To find the indices group partition at each level, we anticipate that something similar to the IANN algorithm in Section \ref{sec:algorithm} can be developed. The main benefit of this extension is that it expands the class of $f$ to which our IANN provides a good approximation (the IANN structure in (\ref{eqn:f}) is a special case of the tree IANN structure with each group of inputs partitioned into a single input and the remaining inputs) and potentially reduces the depth of the tree in the IANN visualization. 


Another potential extension is from the current setting of visualizing black box simulation functions to visualizing general supervised learning models fit to observational data, in order to interpret the effects of the input variables. The main challenge is that, unlike black box simulation functions for which the entire (rectangular) input space is meaningful, the input training data for supervised learning models are often highly correlated. Visualizing black box supervised  learning models using our current IANN approach would require extrapolation to regions of the input space where data are scarce, which would render the interpretations unreliable.

\section*{Supplementary Materials}
\label{sec:supp}
\begin{description}
\item[Supplemental Materials] Several topics will be covered here, including the algorithms for the DASH IANN structure, customized LHD sampling techniques, additional numerical examples, and the proof of Theorem \ref{thm}. (IANN-supplementary.pdf)
\item[Python-package for IANN:] Python-package “IANN” containing code to perform the IANN method described in the article and the related datasets. (iann-codes.zip, zipped codes for IANN)
\end{description}

\section*{Disclosure Statement}
The authors report there are no competing interests to declare.

\section*{Acknowledgements}
This work was funded in part by the Air Force Office of Scientific Research Grant \# FA9550-18-1-0381, which we gratefully acknowledge.

\bibliographystyle{chicago}
\bibliography{ref.bib}

\clearpage

\end{document}